\documentclass[10pt, conference, compsocconf]{IEEEtran}
\IEEEoverridecommandlockouts
\usepackage{cite}
\usepackage{amsmath,amssymb,amsfonts}
\usepackage{algorithmic}
\usepackage{graphicx}
\usepackage{textcomp}
\usepackage{multirow} 
\usepackage{hyperref} 
\renewcommand\figurename{Fig.}

\ifCLASSOPTIONcompsoc
    \usepackage[caption=false, font=normalsize, labelfont=sf, textfont=sf]{subfig}
\else
\usepackage[caption=false, font=footnotesize]{subfig}
\fi

\begin{document}
\title{\Large{\textbf{Dual-encoder Bidirectional Generative Adversarial Networks for Anomaly Detection}}}

\author{\IEEEauthorblockN{Teguh Budianto, Tomohiro Nakai, Kazunori Imoto, Takahiro Takimoto, Kosuke Haruki}
\IEEEauthorblockA{\textit{Corporate Research and Development Center, Toshiba Corporation, Kawasaki, Japan} \\
\{teguh1.budianto, tomohiro.nakai, kazunori.imoto,\\ takahiro.takimoto, kosuke.haruki\}@toshiba.co.jp}
}

\maketitle

\begin{abstract}
Generative adversarial networks (GANs) have shown promise for various problems including anomaly detection. When anomaly detection is performed using GAN models that learn only the features of normal data samples, data that are not similar to normal data are detected as abnormal samples. The present approach is developed by employing a dual-encoder in a bidirectional GAN architecture that is trained simultaneously with a generator and a discriminator network. Through the learning mechanism, the proposed method aims to reduce the problem of bad  cycle consistency, in which  a bidirectional GAN might not be able to reproduce samples with a large difference between normal and abnormal samples. We assume that bad cycle consistency occurs when the method does not preserve enough information of the sample data. We show that our proposed method performs well in capturing the distribution of normal samples, thereby improving anomaly detection on GAN-based models. Experiments are reported in which our method is applied to publicly available datasets, including application to a brain magnetic resonance imaging anomaly detection system. 
\end{abstract}


\section{Introduction} \label{sec:intro}

Anomaly detection is a well-known problem that focuses mainly on finding abnormal data behavior that differs from a normal data distribution. Studies of the anomaly detection problem have benefitted various fields, including health care~\cite{schlegl2017unsupervised}, video surveillance~\cite{sultani2018real} \cite{kiran2018overview}, and image analysis~\cite{haselmann2018anomaly}. Most anomaly detection problems, particularly in high-dimensional image datasets, are defined by separating abnormal samples that are visually different from the data distribution. Separating anomalies from a data distribution can be useful for improving product quality inspections in manufacturing systems~\cite{wang2018deep}, detecting brain tumors in medical images~\cite{chen2018unsupervised}, and detecting anomalous objects in video surveillance~\cite{sultani2018real} \cite{kiran2018overview}.

In real-world applications, there is a strong need for anomaly detection techniques that are able to handle distributions of complex high-dimensional data. In terms of data complexity, however, conventional anomaly detection methods are unsuitable for solving the aforementioned problems~\cite{kim2019forward}\cite{ngo2019fence}. Usually, only a small number of anomalous samples are available, which leads to collection of an imbalanced data sample. This phenomenon has led researchers to propose learning approaches in semi-supervised and unsupervised settings, such as image reconstruction-based anomaly detection systems.

Anomaly detection methods based on generative adversarial networks~(GANs) have shown promising performance in capturing the distributions of high-dimensional and complex data~\cite{schlegl2017unsupervised}\cite{zenati2018efficient}. In particular, Zenati et al.~\cite{zenati2018efficient} proposed an anomaly detection framework employing a bidirectional GAN~(BiGAN)\cite{dumoulin2017adversarially}\cite{donahue2017adversarial} that simultaneously learns an encoder, a generator, and a discriminator during training.

\begin{figure}
    \centerline{
    \includegraphics[scale=0.38]{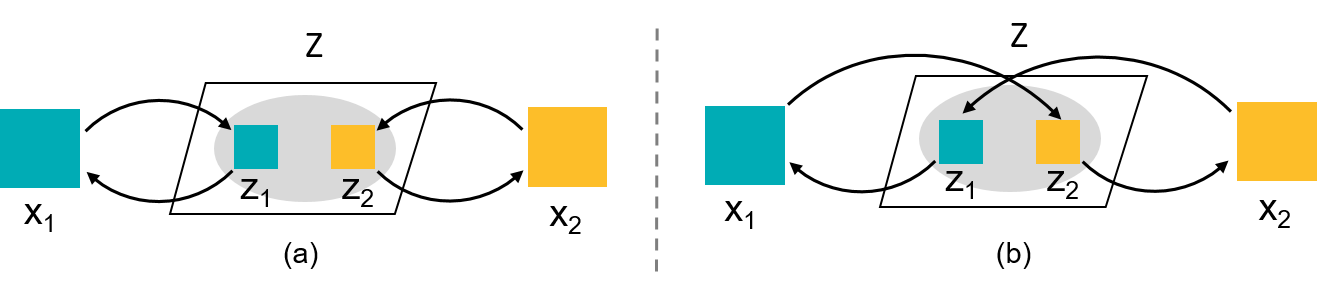}}
    \caption{Cycle consistency showing samples in data space and latent space $Z$. All samples in data space are represented by $x$. The data samples $x_1$ and $x_2$ are in the normal data distribution, and the predefined latent variables $z_1$ and $z_2$ are in latent space $Z$ distributed in a circle. (a) Good cycle consistency that reproduces input samples consistently. (b) Bad cycle-consistency in which point $x_1$ maps to $z_2$ and reproduces $x_2$ instead of the original point $x_1$. The depicted concept is introduced in~\cite{kim2019forward}.}
    \label{fig:cycle-consistent}
\end{figure}

BiGAN is trained through bidirectional adversarial learning in which an encoder and a generator network are used to generate data both in data and latent space~\cite{kim2019forward}. Working similarly to an autoencoder in data reconstruction, BiGANs generate normal and abnormal samples similar to the normal samples in order to measure the abnormality through reconstruction error. Data reproduction quality in BiGANs has shown limitations in normal sample reconstruction, resulting in high reconstruction error for normal samples. This could significantly degrade the anomaly detection performance~\cite{dumoulin2017adversarially}\cite{donahue2017adversarial}. This creates an insufficient difference between samples. The condition where a model cannot reproduce samples and gives large reconstruction error is called bad cycle consistency~\cite{zhu2017unpaired}. Fig.~\ref{fig:cycle-consistent} shows a conceptual image of cycle consistency.

This study assumes that bad cycle consistency can occur as a result of a model not preserving enough information of the input image. The proposed method introduces \textit{preserved information learning} using a dual-encoder in BiGAN. The aim of preserved information learning in dual-encoder BiGAN is to significantly reduce the bad cycle consistency of GAN-based architectures. In this paper, a model employing dual-encoder BiGAN is proposed for addressing bad cycle consistency and improving anomaly detection. Toward these ends, the main goals of this paper are as follows.

\begin{itemize}
  \item To propose a GAN-based anomaly detection technique for reducing bad cycle consistency.
  \item To evaluate the proposed method on several publicly available datasets and demonstrate its performance compared with state-of-the-art methods.
\end{itemize}

\section{Related Work} \label{sec:related-work}

A complete review of deep learning technology for anomaly detection along with its application across various domains is comprehensively  explained in~\cite{chalapathy2019deep}. Examples of existing methods include autoencoder-based anomaly detection, such as denoising autoencoder~\cite{vincent2008extracting}, robust deep autoencoder~\cite{zhou2017anomaly}, and variational autoencoder~\cite{an2015variational}. Autoencoder-based methods usually detect anomalies by measuring the difference between an original sample $x$ and its reconstruction $x'$ as $\|x-x'\|$.

Recent anomaly detection methods employing GANs can handle the presence of anomalous samples~\cite{mattia2019a}. GANs train two different networks simultaneously through a minimax game in which one network is a generator $(G)$ that learns to generate data (\textit{e.g.} images) and minimize error, and the other network is a discriminator $(D)$ that aims to distinguish the generated data by $G$ from the real data distribution. The first work that used GANs for anomaly detection, called AnoGAN, was proposed by Schleg et al.~\cite{schlegl2017unsupervised}. AnoGAN is trained using only normal samples to learn a mapping of the latent space representation. During the testing period, the latent vector that best reconstructed the test image is then searched through the latent space representation. The anomaly score in AnoGAN is defined using a combination of reconstruction loss and the difference between the intermediate discriminator feature representation of a test image and its reconstruction. Furthermore, GANomaly~\cite{akcay2018ganomaly} was proposed which uses conditional GANs that jointly learn the generation of a high-dimensional image and the inference of the latent space. GANomaly frameworks consist of encoder-decoder-encoder sub-networks in the generator and a discriminator network. GANomaly defines a new anomaly score as a combination of three loss functions, namely, feature matching loss, reconstruction loss, and encoding loss. A more recent method, called Fence GAN~\cite{ngo2019fence}, aims to generate data lying on the boundary of the normal data distribution by proposing the use of encirclement loss for the GAN loss function. In Fence GAN, the anomaly score is calculated directly using the score from the discriminator. Sabokrou et  al.~\cite{sabokrou2018adversarially} proposed a method that is mainly composed of two networks that are trained adversarially in an unsupervised learning setting. One of the networks in this architecture learns to refine noisy input images, while the other is responsible for separating normal and abnormal sample images.

\section{Background}
\subsection{Generative Adversarial Networks}
GANs consist of two networks for learning data generation. One network is a generator $G$ that learns to generate data and to minimize error, while the other is a discriminator $D$ that aims to distinguish generated data by $G$ from the real data distribution. Both network $G$ and $D$ are trained simultaneously to minimize their loss through a two-player min-max game, formulated as

\begin{equation} \label{eq:gan}
\min_{G} \max_D \mathbb{E}\big[\log D(x) + \log(1-D(G(z)))\big]
\end{equation}

GANs are also used to solve anomaly detection problems and defined as AnoGAN~\cite{schlegl2017unsupervised}.

\subsection{Efficient GAN-based Anomaly Detection }
BiGAN learns an encoder $E$ that maps input samples $x$ to a latent representation $z$, and a generator $G$ and a discriminator $D$ are also trained at the same time. Unlike the original GAN, the discriminator $D$ in BiGAN considers not only the input $x$, but also its respective latent variable $z$. In particular, the BiGAN training objective is defined as

\begin{equation} \label{eq:bigan}
\min_{G,E} \max_D \mathbb{E}\big[\log D(x, E(x)) + \log(1-D(G(z),z))\big]
\end{equation}

To generate $G, D$, and $E$, a model is trained only using normal samples and the anomaly score function $A(x)$ as in~\cite{schlegl2017unsupervised} is used to measure the level of abnormality. The anomaly score is based on the convex combination of reconstruction loss $L_G$ and discriminator loss $L_D$.

\begin{equation} \label{eq:ax-score}
A(x) = \alpha L_G(x) + (1-\alpha)L_D(x)
\end{equation}

\begin{figure}[h]
    \centering
    \includegraphics[scale=0.48]{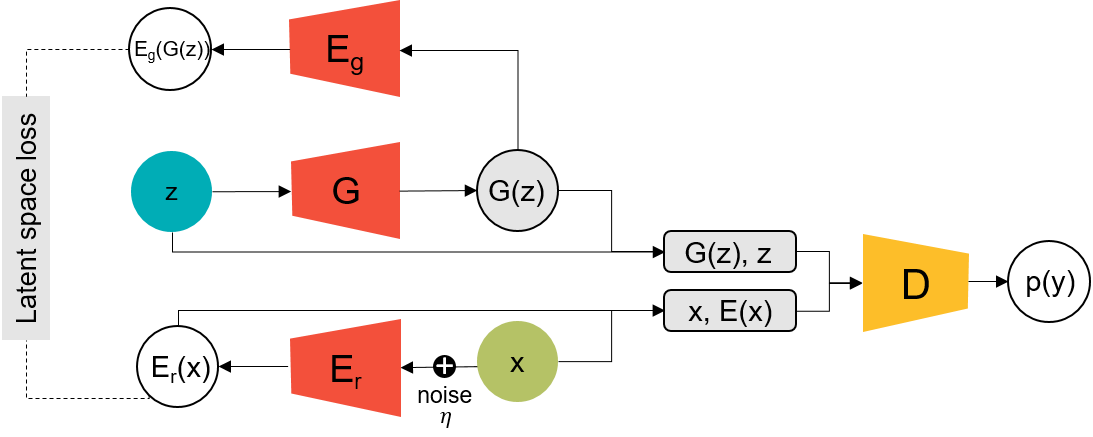}
    \caption{Proposed method: Original method dual-encoder BiGAN architecture. For simplicity, the other losses in the proposed method are omitted from the illustration.}
    \label{fig:dual-encoder-bigan-ori}
\end{figure}

\section{Proposed Method} \label{sec:method}
\subsection{Dual-Encoder BiGAN} 

In this study, a GAN-based anomaly detection approach is proposed for handling bad cycle consistency. BiGAN forces abnormal samples to be reproduced within a normal distribution, but either the normal or abnormal samples suffer from the problem of poor reconstruction of inlier samples, making it difficult for GAN-based methods to detect outlier samples precisely. We assume that bad cycle consistency might occur when a model is unable to preserve enough information of the input image. The proposed method introduces preserved information learning employing a dual-encoder BiGAN architecture (Fig.~\ref{fig:dual-encoder-bigan-ori}). In this case, $\eta$ as depicted in Fig.~\ref{fig:dual-encoder-bigan-ori} is a Gaussian noise that is added to input sample $x$ in order to make the proposed method more robust against corrupted samples. Furthermore, the preserved information learning in a dual-encoder BiGAN uses cycle consistency loss and latent space variable loss. In Fig.~\ref{fig:dual-encoder-bigan-ori}, $p(y)$ represents the probability that the joint input of a sample and latent variable to discriminator $D$ comes from a real or fake sample.

The proposed method optimizes the generator $G$ by prioritizing cycle consistency. The goal is to be able to generate an image that can be reconstructed back to its original source. It is expected that the proposed method can overcome the main problems by prioritizing the cycle consistency loss. Furthermore, an evaluation is performed on publicly available image datasets, including a real-world medical image dataset (Section~\ref{sec:discussion}).

\begin{figure}
    \centering
    \includegraphics[scale=0.42]{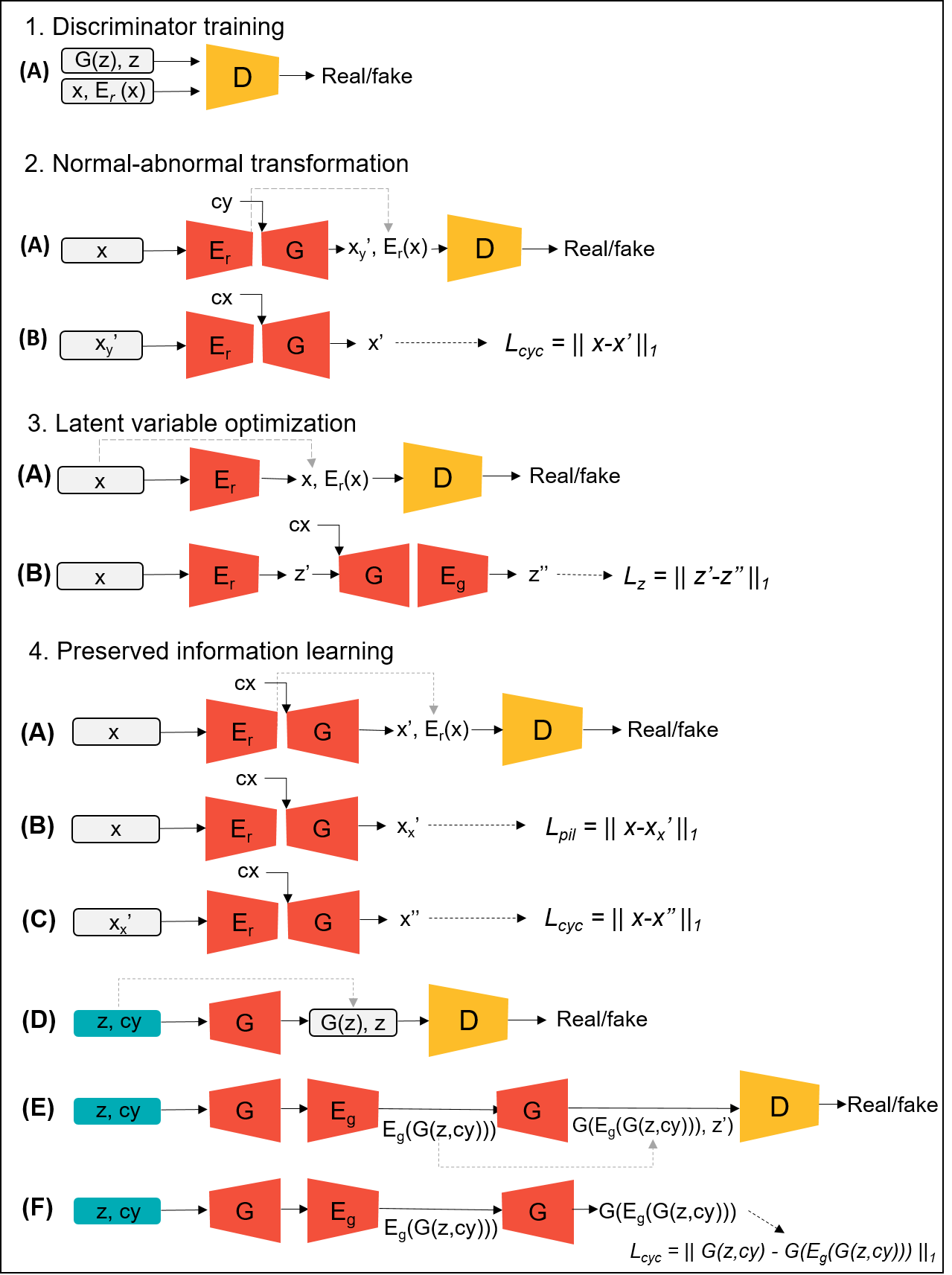}
    \caption{Training mechanism for dual-encoder BiGAN.}
    \label{fig:training-mechanism}
\end{figure}

\subsection{Preserved Information Learning}
Fig.~\ref{fig:training-mechanism} shows the training mechanism for training all networks in the proposed method,. This training mechanism is implemented simultaneously in order to train each of the networks in the dual-encoder BiGAN architecture. The concept behind the complete training scheme of the dual-encoder BiGAN is inspired by image-to-image translation methods~\cite{choi2018stargan}\cite{siddiquee2019learning}. In image-to-image translation, the image transformation is done through different-domain or same-domain transformation. In contrast, the proposed method does not use any domain information, but instead uses single-class input. 

The complete training scheme as shown in Fig.~\ref{fig:training-mechanism} is as follows. 
\begin{itemize}

\item{Discriminator $D$ is trained adversarially to separate real/fake images (see Fig.~\ref{fig:training-mechanism}-1). As in BiGAN, the discriminator input is a pair of samples in image space and its respective latent variable, both for real sample $x$ and fake/reconstructed sample $G(z)$.}

\item{In preserved information learning, the networks are conditioned using the target variable $c$. The target variable $c$ is an extra information input provided to  the conditioning function by feeding the real target $c_x$ or random target $c_y$ as an additional input layer to the networks. Both $c_x$ and $c_y$ are configured to control the generation of a sample corresponding to the source, whether it is from real data sample $x$ or from generated sample.}

\item{As shown in Fig.~\ref{fig:training-mechanism}-2(A-B), a real sample $x$ and random target $c_y$ are regenerated through $G(E_r(x),c_y)$ as generated sample $x_y'$, which is then reconstructed back to $x'$ in order to measure the loss of cycle consistency of sample $x$. This procedure is called normal-abnormal transformation because it employs a random target $c_y$ (not a real normal label) that is uniformly distributed as input to encoder $E_r$ during training (see Fig.~\ref{fig:training-mechanism}-2 (A)). The input of $c_x$ is the real label used for real normal sample $x$.}

\item{The present architecture employs a dual-encoder in which the second encoder $E_g$ is proposed in order to optimize the distance between a real sample latent variable and reconstructed latent variable in latent space. Fig.~\ref{fig:training-mechanism}-3 shows the latent variable optimization in the proposed method. In particular, the output from discriminator $D$ is also used to update the encoder $E_r$ that appears in Eq.~\ref{eq:en1-loss}, as shown in Fig.~\ref{fig:training-mechanism}-3(A). In comparison with the bottom of Fig.~\ref{fig:training-mechanism}-1(A) where measurement is performed to update the discriminator $D$, in Fig.~\ref{fig:training-mechanism}-3(A) it is used to update the encoder $E_r$. Furthermore, the effect of latent space optimization on dual-encoder BiGAN is shown in Fig.~\ref{fig:mnist-proposed-non-tr} after BiGAN (Fig.~\ref{fig:mnist-egbad}) with additional encoder $E_g$.}

\item{Preserved information learning (Fig.~\ref{fig:training-mechanism}-4) employs input from both random variable $z$ and latent variable $E_r(x)$ into generator $G$. As this is expected to reduce bad cycle consistency, we additionally support this process by prioritizing cycle consistency loss in generator $G$. As a result, the generator learns to enrich its ability to generate various samples and is also pushed to reconstruct input samples from $E_r(x)$ similarly to an auto-encoder. The improvement achieved by using this procedure can be seen in Fig.~\ref{fig:mnist-cycle-consistency-evidence}, which shows that the proposed method with preserved information learning reduced bad cycle consistency in the MNIST anomaly dataset.
The generator is trained through a preserved information learning scheme as shown in Fig.~\ref{fig:training-mechanism}-4 with the assistance of cycle consistency prioritization. The encoder employs an input image $x$ and its target $c$. For any generated or reconstructed image, target $c_y$ is used, where $c_y$ is defined as a noisy random target provided to the generator for learning features from non-normal image input. }
\end{itemize}

By learning only normal features, the reconstruction in the proposed method is closed to normal samples. This mechanism therefore provides two advantages: (1) generator $G$ reconstructs the image to normal features; and (2) discriminator $D$ is able to measure any input $x$ as normal or abnormal. To realize these training mechanisms, dual-encoder BiGAN losses are defined as follows.

\begin{figure}
    \centering
    \includegraphics[scale=0.28]{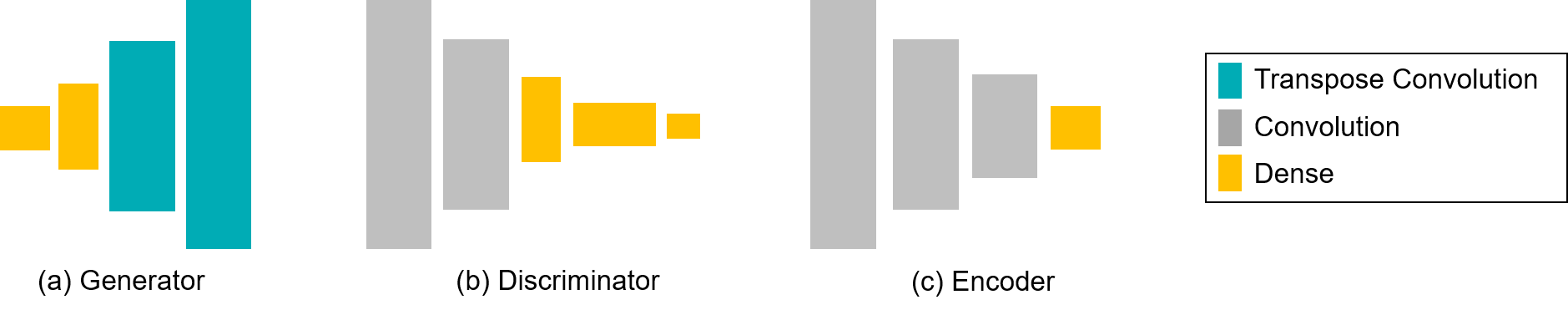}
    \caption{Networks structure of (a) generator $G$, (b) discriminator $D$, and (c) encoders $E_r$ and $E_g$. }
    \label{fig:network-layers}
\end{figure}

\subsubsection{Adversarial loss}
In the proposed method, the generator learns to preserve information from the input images. To ensure that generator $G$ is able to judge a real/fake latent variable $z$, the generator also receives signals from both input images and random latent variables. As mentioned earlier, the proposed method prioritizes cycle consistency, which makes it important to learn the relationship between image space and latent space. The adversarial loss of discriminator $D$ is modified as follows.

\begin{equation} \label{eq:adversarial-loss}
\mathcal{L}_{adv}^{D} = \sum_{c \in \lbrace c_y, c_x \rbrace} \mathbb{E} \big[ \log(1-D(G(z,c),z)) + \log D(x,E_r(x)) \big]
\end{equation}

The adversarial loss of generator $G$ is modified by adding the loss information of encoder $E_g$

\begin{equation} \label{eq:gen-adversarial-loss}
\begin{aligned} 
\mathcal{L}_{adv}^{G} = & \Bigg( \sum_{c \in \lbrace c_y, c_x \rbrace} \mathbb{E} \big[ \log(1-D(G(z,c),z)) \big] \Bigg) + \\
&  \mathbb{E} \big[ \log(1-D(G(E_g(G(z,c_y))), E_g(G(z,c_y)))) \big]
\end{aligned}
\end{equation}

\subsubsection{Prioritized cycle consistency loss}
In dual-encoder BiGAN, the generator is optimized by prioritizing cycle consistency loss. Cycle consistency loss helps the generator to preserve enough information for reconstructing the generated image back to its original.

\begin{equation}\label{eq:cycle-loss}
\begin{aligned} 
\mathcal{L}_{cyc} = & \mathbb{E}_{x,c_x,c_y} \Big(
\Big[ \Vert G(E_r(G(E_r(x),c_y)), c_x) - x \Vert_1 \Big] \\
& + \Big[ \Vert G(E_r(G(E_r(x),c_x)), c_x) - x \Vert_1 \Big] \\
& + \Big[ \Vert G(E_g(G(z)),c_y) - G(z,c_y) \Vert_1 \Big] \Big)
\end{aligned}
\end{equation}

\subsubsection{Preserved information loss}

We modify the identity loss from~\cite{siddiquee2019learning} as \textit{preserved information loss} to penalize generator $G$ when learning real input images.

\begin{equation} \label{eq:preserved-loss}
\mathcal{L}_{pil} = \begin{cases}
0 , & c = c_y \\
\mathbb{E}_{x,c} = \big[ \Vert G(E_r(x),c) - x \Vert_1 \big] , & c = c_x
\end{cases}
\end{equation}

\subsubsection{Latent space loss}
We introduce a second encoder $E_g$ for minimizing the distance between $z$ and its latent reconstruction in latent space.

\begin{equation} \label{eq:latent-z-loss}
\mathcal{L}_z = \big[ \Vert E_g(G(E_r(x),c_x)) - E_r(x) \Vert_1 \big]
\end{equation}

where $\mathcal{L}_z$ is only introduced for the real input $x$ while the second term in~\ref{eq:preserved-loss} is penalized through a priority parameter for cycle consistency.

\subsubsection{Full objective}

Total optimization of discriminator $D$, generator $G$, encoder $E_r$ and $E_g$ in Dual-encoder BiGAN is as follows.

\begin{equation} \label{eq:dis-loss}
\mathcal{L}_D = \mathcal{L}_{adv}^{D}
\end{equation}

\begin{equation} \label{eq:gen-loss}
\mathcal{L}_G = \mathcal{L}_{adv}^{G} + \lambda_{cyc}\mathcal{L}_{cyc} + \mathcal{L}_{pil}
\end{equation}

\begin{equation} \label{eq:en1-loss}
\mathcal{L}_{E_r} = \mathbb{E} \big[\log D(x, E_r(x))\big] + \mathbb{E}\big[ \Vert G(E_r(x),c_x) - x \Vert_1 \big]
\end{equation}

\begin{equation} \label{eq:en2-loss}
\mathcal{L}_{E_g} = \mathcal{L}_z
\end{equation}

Here, $\lambda_{cyc}$ is a priority parameter of cycle consistency for generator $G$. In our experiments, we set $\lambda_{cyc}$ = 0.1.

\subsection{Anomaly Score}

The proposed method is trained using only normal samples and employs a preserved information learning mechanism (Fig.~\ref{fig:training-mechanism}-4). The anomaly score is defined as in Equation~\ref{eq:ax-score} where we use $\alpha$ = 0.1, which has been found empirically through experiments in~\cite{schlegl2017unsupervised}. We also find that the discriminator score $D(x)$ can also be employed as an alternative anomaly score. In the evaluation phase, real target $c_x$ is substituted for only normal targets because the trained model is familiar with only normal input samples.

\section{Experiments} \label{sec:discussion}
To evaluate the proposed method, extensive experiments were conducted on publicly available datasets. In all experiments using the unsupervised learning setup, the proposed method used only normal samples to train the models. A common practical performance metric, the area under the receiver operating characteristics curve (AUROC) is used to measure the quality of interchangeability of the given scoring of the methods proposed in this study.

\subsection{Architecture}
In our experiments, we follow the network structure for all generator, discriminator, and encoders introduced in BiGAN~\cite{zenati2018efficient} when evaluating the MNIST dataset. There are only minimal differences between the structures of our networks and the original BiGAN; these differences are due to the join input of target $c$ and the latent space variable. In particular, Fig.~\ref{fig:network-layers} shows the number of layers used in the proposed method, which consist mainly of Dense, Convolution, and Convolution Transpose layers. For fair comparison, we use the same architecture for the baselines with minimal differences between each of the architectures. 

\begin{figure}
    \centering
    \includegraphics[scale=0.45]{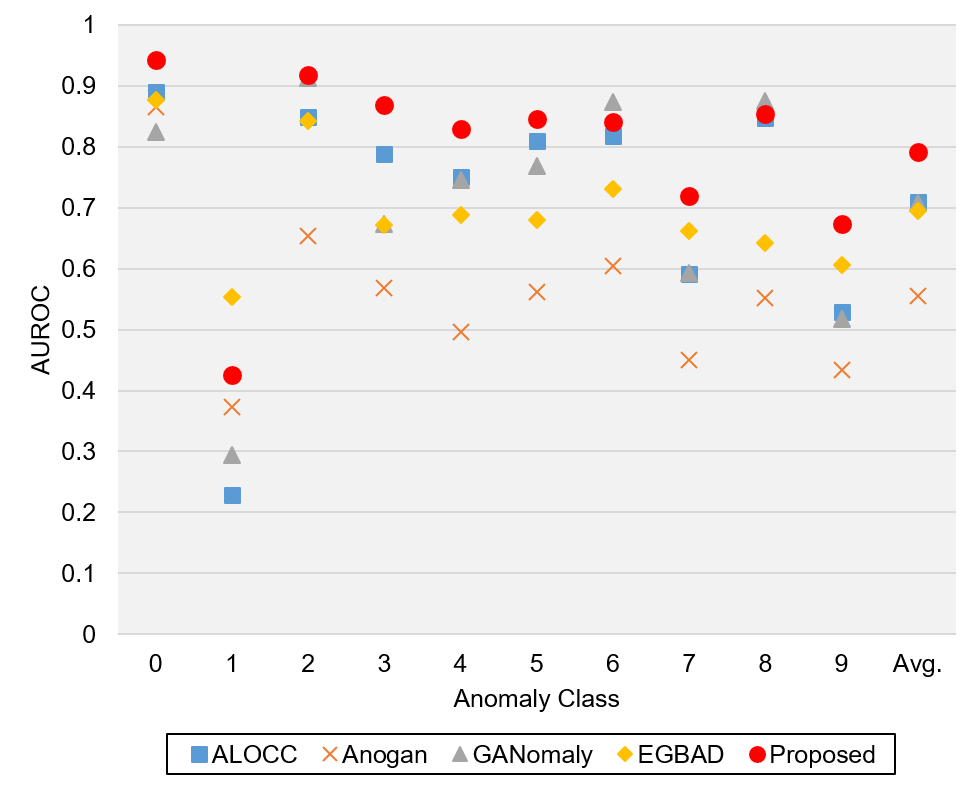}
    \caption{AUROC results on MNIST dataset.}
    \label{fig:mnist-results-vs-sota}
\end{figure}

\begin{figure}
    \centering
    \includegraphics[scale=0.45]{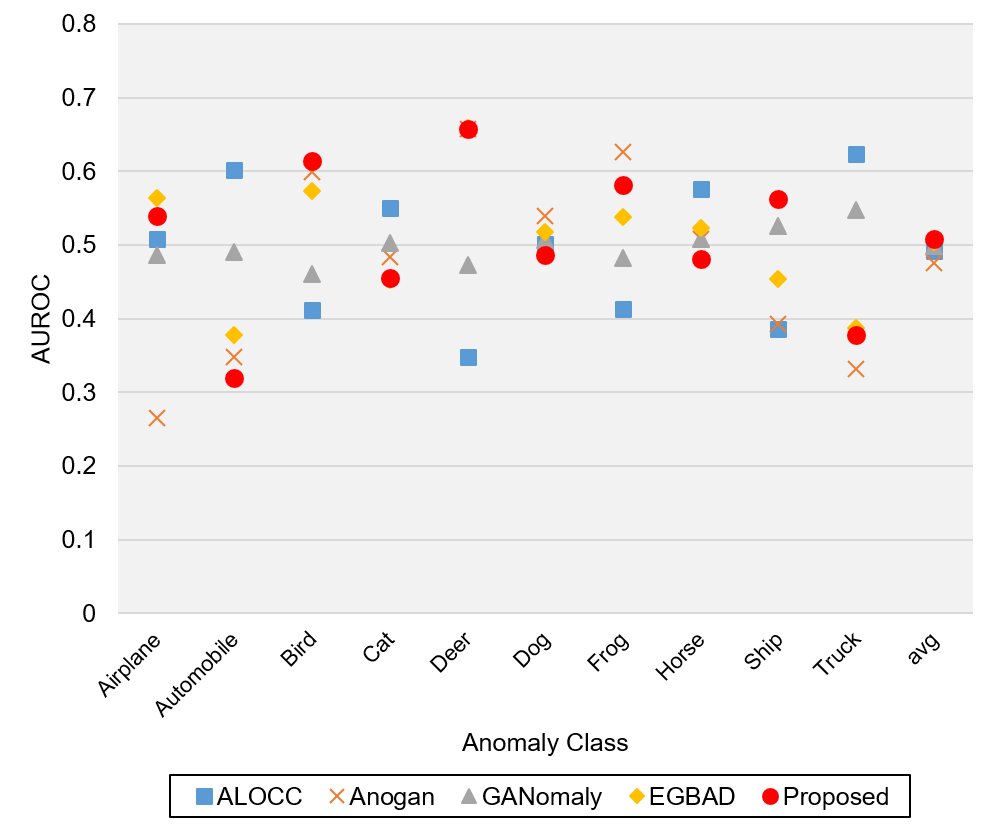}
    \caption{AUROC results on the CIFAR10 dataset}
    \label{fig:cifar10-results-vs-sota}
\end{figure}

\subsection{Datasets}
To evaluate the proposed method, experiments were conducted on three publicly available datasets. The following is a brief overview of each dataset used for evaluation of the proposed method.

\subsubsection{MNIST}
The MNIST dataset contains handwritten digits which are usually used for early stage model evaluation. The data are split between 60,000 samples in the training set and 10,000 samples in the test set. Within each set, there are $28\times 28$ pixel grayscale images with a total of 10 output classes representing the ten digits from 0 to 9. The evaluation presented in this work was conducted on each class in which, at any time, only one class is considered as the anomaly class, and the remaining nine classes are considered together as the normal class. This means that only the normal examples in the training set were used to train the evaluated models while the training data of the class considered abnormal were ignored. The test procedure was applied to a test set that had not been seen by the trained model.

\subsubsection{CIFAR10}
The CIFAR10 dataset~\cite{krizhevsky2009learning} contains natural color $32\times32$ pixel images. The dataset is split into images associated with labels representing objects of 10 classes of natural images, such as image of "airplane", "automobiles", and "dog". As with the MNIST dataset, we configured the dataset for each class by using only one class as the anomaly object and trained the models only using normal data samples from the rest of the classes.

\subsubsection{BRATS 2013}
The BRATS 2013 dataset~\cite{menze2014multimodal}\cite{kistler2013virtual} consists of synthetic and real images. Each image is divided into healthy samples and tumor positive samples with high-grade gliomas (HGs) and low-grade gliomas (LGs). There are 25 patients with both synthetic HG and LG images and 20 patients with real HG and 10 patients with real LG images. In this case, we are not particularly trying to segment the tumors, but rather trying to predict the separation of whether an image contains a tumor, which represents an abnormality in the sample.

\begin{figure}
    \centering
    \includegraphics[scale=0.53]{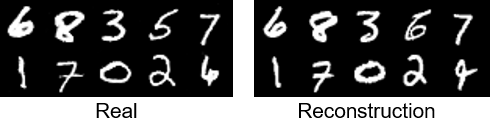}
    \caption{Image reconstruction of anomaly digit 1. The proposed method completely reconstructs normal and abnormal samples. Fundamentally, we expect the proposed method to be unable to reconstruct any abnormal samples. This phenomenon could have the effect that the anomaly score $A(x)$ is not able to separate normal and abnormal samples. }
    \label{fig:mnist-anomaly-digit-1}
\end{figure}

\begin{figure*}[ht]
    \centering
    	\subfloat[Real\label{fig:mnist-real}]{%
        \includegraphics[scale=0.55]{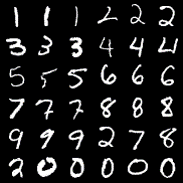}}
	~
    	\subfloat[EGBAD~\cite{zenati2018efficient}\label{fig:mnist-egbad}]{%
        \includegraphics[scale=0.55]{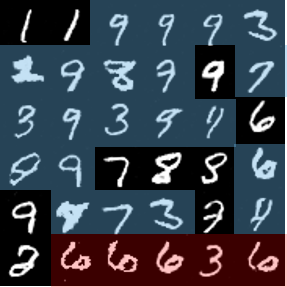}}
	~
    	\subfloat[Proposed (without preserved information learning)\label{fig:mnist-proposed-non-tr}]{%
        \includegraphics[scale=0.55]{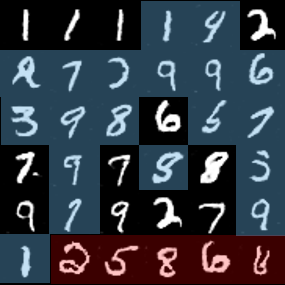}}
    ~    
    	\subfloat[Proposed\label{fig:mnist-proposed}]{%
        \includegraphics[scale=0.55]{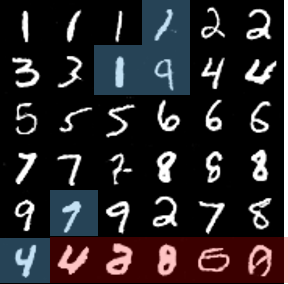}}

    \caption{Evaluation results for abnormal digits 0. Digits in blue boxes indicate models that were unable to reconstruct normal samples due to bad cycle consistency, while digits in red boxes indicate models that were unable to reconstruct abnormal images. (a) Real input images. (b) Reconstructed images by BiGAN/EGBAD. (c) Reconstructed images by proposed architecture dual-encoder BiGAN trained adversarially without preserved information learning. (d) Proposed method.}
    \label{fig:mnist-cycle-consistency-evidence}
\end{figure*} 

\subsection{Results}

\subsubsection{Application to MNIST}
Fig.~\ref{fig:mnist-results-vs-sota} shows the AUROC results obtained using the MNIST dataset, where the x-axis represents anomalous classes and the average overall performance. As shown in Fig.~\ref{fig:mnist-results-vs-sota}, the proposed method outperforms the state-of-the-art methods on the majority of abnormal class digits. Interestingly, the proposed method performs badly against only anomaly digit 1. This may occur due to the proposed method completely reconstructing all normal and abnormal samples as shown in Fig.~\ref{fig:mnist-anomaly-digit-1}. This causes the anomaly score for anomaly digit 1 to be very close to that of the digits in the normal class.

Fig.~\ref{fig:mnist-cycle-consistency-evidence} shows the reconstruction images from generator $G$ of the BiGAN-based methods. In the figures, digit class 0 was selected as the anomaly class. The original BiGAN evaluated in \cite{zenati2018efficient} (EGBAD) shows the condition in which normal samples were generated the same as other digits among normal samples. This evidence shows that BiGAN~(EGBAD) suffers from bad cycle consistency, which may cause high reconstruction error of normal samples (Fig.~\ref{fig:mnist-egbad}). The proposed method overcomes this shortcoming by employing a dual-encoder in the BiGAN architecture. Fig.~\ref{fig:mnist-proposed} shows the results of preserved information learning in dual-encoder BiGAN. This significantly reduces the bad cycle consistency, thereby leading to improved detection performance. Preserved information learning is important for achieving the maximum capability of dual-encoder BiGAN. Based on empirical observations, dual-encoder BiGAN is not expected to reach its best performance by only adversarial training with the help of latent space optimization (Fig.~\ref{fig:mnist-proposed-non-tr}). This shows the benefit of applying dual-encoder to BiGAN with a complete preserved information learning mechanism.

\begin{table}
  \begin{center}
    \caption{Performance on the BRATS 2013 dataset according to AUROC metric.}
    \label{tab:brats-results-auroc}
    \begin{tabular}{l|r} 
      	Methods & AUROC \\
      	\hline \hline
      	AnoGAN & 0.340279 \\
      	ALOCC & 0.620300 \\
      	GANomaly & 0.845130 \\
		EGBAD + A(x) & 0.286299 \\
		EGBAD + D(x) & 0.786707 \\
		Proposed + A(x) & 0.632038 \\
		Proposed + D(x)	& \textbf{0.861377} \\
		\hline
    \end{tabular}
  \end{center}
\end{table}

\subsubsection{Application to CIFAR10}
Fig.~\ref{fig:cifar10-results-vs-sota} shows a comparison of the performance results for the proposed method compared with the baselines for the CIFAR10 anomaly dataset. Since the images in CIFAR10 contain natural images which have more complex visual structures compared with the MNIST images, it shows that all GAN-based anomaly detection, including ours, offer fair results without providing high performance in the separation of normal and abnormal samples. On average, the proposed method is quite competitive against the baselines, particularly for the anomaly class 'deer'. The CIFAR10 dataset clearly provides a different level of difficulty and is a challenging problem
for the anomaly detection task. Fig.~\ref{fig:cifar10-latent-size-comparison} shows the reconstruction image of anomaly class 'airplane' for different latent variable sizes.

\begin{table}
  \begin{center}
    \caption{Performance of modified versions of the proposed method on BRATS 2013 when changing latent variable size.}
    \label{tab:brats-ablation-study}
    \begin{tabular}{l|c|r} 
		Modification & z & AUROC \\
		\hline \hline
		Proposed + A(x)	& \multirow{2}{*}{20} & 0.632038 \\
		Proposed + D(x)	&    & 0.861377 \\ \hline
		Proposed + A(x) & \multirow{2}{*}{50} & 0.743657 \\
		Proposed + D(x)	&    & 0.847963 \\ \hline
		Proposed + A(x)	& \multirow{2}{*}{100} & 0.634741 \\
		Proposed + D(x)	&    & \textbf{0.926704} \\
		\hline
    \end{tabular}
  \end{center}
\end{table}

\subsubsection{Application to brain magnetic resonance imaging anomaly detection}

\begin{figure}
    \centering
    	\subfloat[Proposed\label{fig:brats-reconstruction-proposed}]{%
        \includegraphics[scale=0.40]{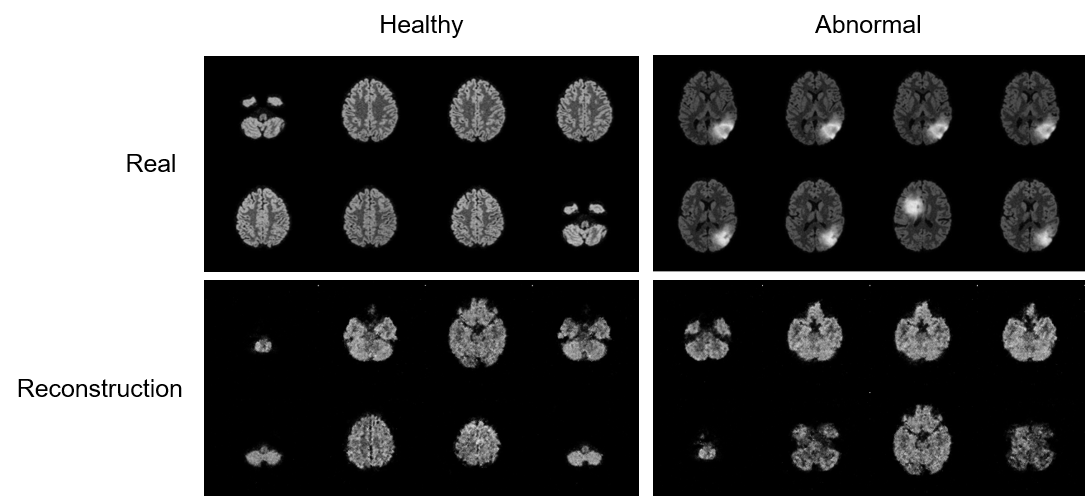}}
 	
 		\subfloat[EGBAD\label{figbrats-reconstruction-egbad}]{%
        \includegraphics[scale=0.40]{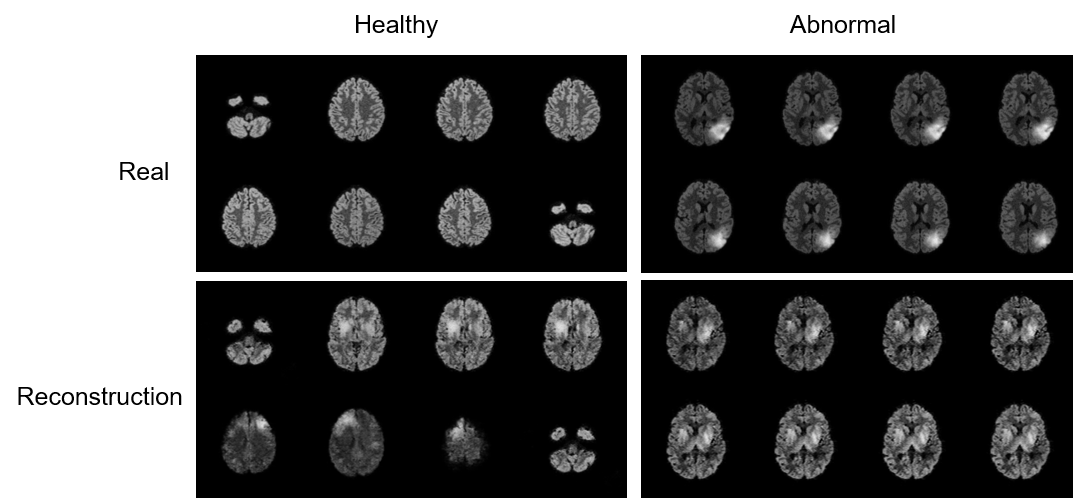}}

    \caption{Reconstruction of healthy and abnormal samples by (a) the proposed method and (b) EGBAD.}
    \label{fig:brats-reconstruction}
\end{figure}

We employed the proposed method and EGBAD to the brain magnetic resonance imaging problem domain, specifically the BRATS 2013 dataset, and used the anomaly score $A(x)$ and output of discriminator $D(x)$ as anomaly scoring. Proposed model with anomaly scoring A(x) and D(x) is presented as \textit{proposed + A(x)} and \textit{proposed + D(x)}, respectively. This configuration is also applied for EGBAD. The performance results obtained by the proposed method are shown in Table~\ref{tab:brats-results-auroc}. Overall, we see that the proposed method gave the highest performance on the BRATS 2013 dataset as indicated by the AUROC score  (AUROC: 0.861377) and was competitive against GANomaly. Fig.~\ref{fig:brats-reconstruction} shows a comparison of the reconstruction with the EGBAD method for BRATS 2013. The healthy image reconstruction by the proposed method seems worse than that by EGBAD. Since our best reconstruction is obtain by D(x), the reconstruction error is not very important for helping us find the normal and abnormal samples in BRATS 2013. In addition, there is a tendency for EGBAD to fall into mode collapse by qualitatively examining the reconstruction of both healthy and abnormal samples.

\subsection{Modified Versions of Dual-encoder BiGAN}

This section presents further studies for developing modified methods to dual-encoder BiGAN. In the evaluation on a relatively simple dataset (MNIST), dual-encoder BiGAN showed strong performance against most abnormal classes. However, the proposed method may not be able to achieve the best results due to differences in data complexity and characteristics, as shown in the experimental results for CIFAR10. The following are proposed modifications of the proposed method to achieve improved performance through selection of the architecture and training mechanism.

\subsubsection{Simple vs. complete training scheme}
\begin{figure}
    \centering
    \includegraphics[scale=0.52]{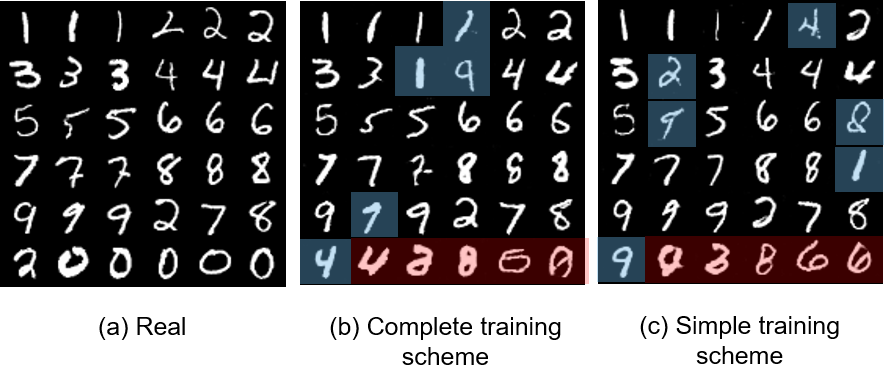}
    \caption{Cycle consistency of different training schemes on MNIST anomaly digit 0.}
    \label{fig:mnist-training-scheme}
\end{figure}

\begin{figure}
    \centering
    \includegraphics[scale=0.44]{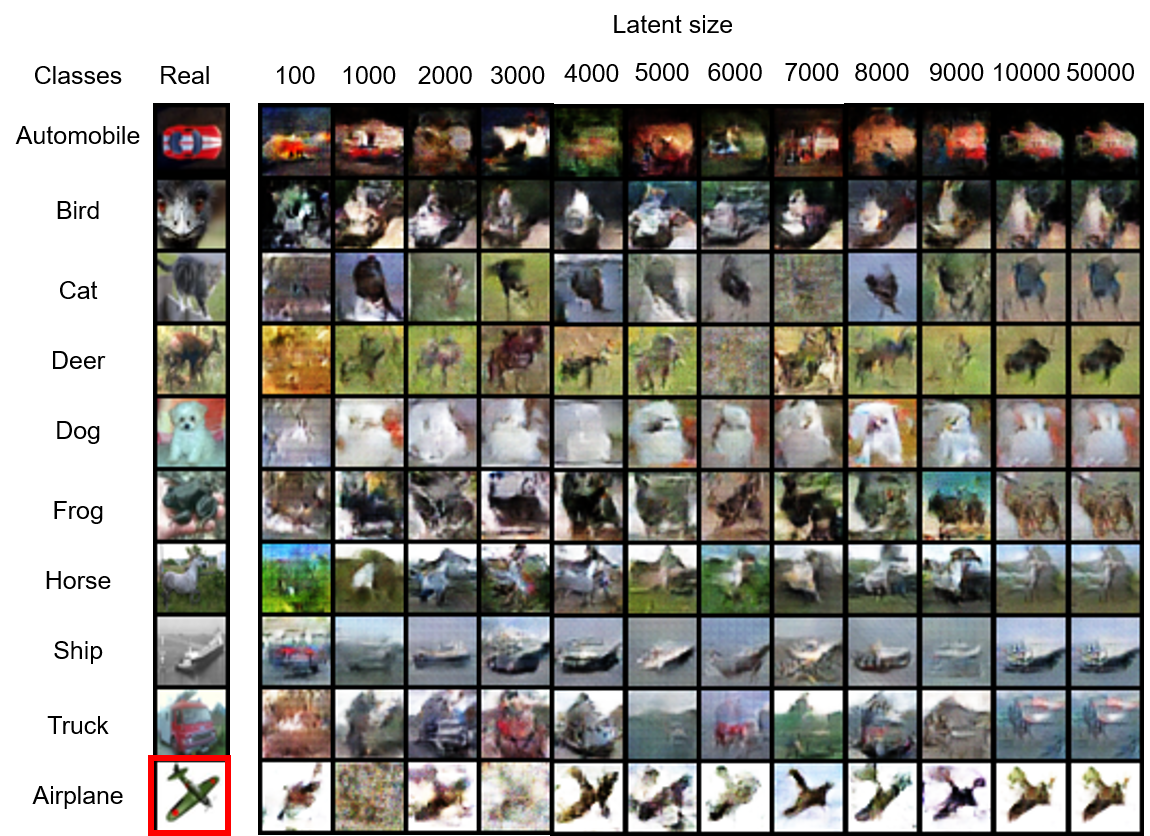}
    \caption{Reconstruction image for different latent variable sizes on CIFAR10}
    \label{fig:cifar10-latent-size-comparison}
\end{figure}

The proposed method offers competitive performance through its complete training scheme. Here we propose simple versions of the proposed method: (1) without providing a random target to the generator $G$; and (2) omitting step 2 (A-B) shown in Fig.~\ref{fig:training-mechanism} from the training mechanism since it is not needed when the random target is provided to generator $G$. The simple training scheme performs well enough on the MNIST dataset that it does not completely suffer from bad cycle consistency (see Fig.~\ref{fig:mnist-training-scheme}). While it does not show major degradation on the non-complicated dataset, complete reconstruction of natural image datasets requires further improvement as in the case of  CIFAR10 (see the comparison of a real sample and reconstruction in Fig.~\ref{fig:cifar10-latent-size-comparison}).

\subsubsection{Effect of latent variable size}

This section considers the effect of latent variable size on both encoder $E_r(x)$ and $E_g(x)$ and their random variables. These variables share the same size and contain the information required by the generator to regenerate an image. Our assumption is the size of $z$ could be critical for deciding the precision of information required by the generator to translate $z$ into image space. 

In addition, changing the latent variable size is another modification for improving the anomaly detection performance presented in Table~\ref{tab:brats-ablation-study}. This improves the AUROC performance of the proposed method from 0.861377 to 0.926704 on BRATS 2013.

\subsubsection{Latent space projection}
\begin{figure}
    \centering
		\subfloat[Encoder $E_r$\label{fig:mnist-projection-first-enc}]{%
        \includegraphics[width=1.8in]{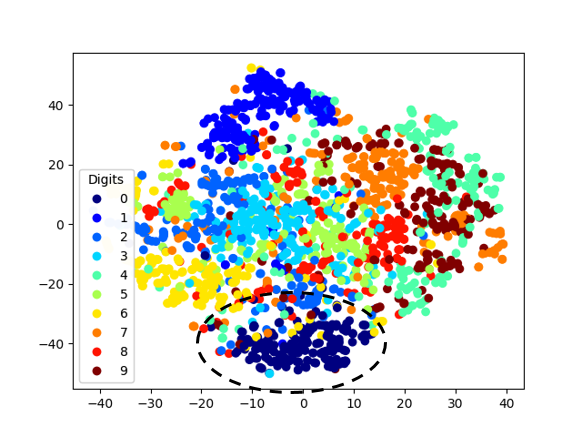}}
    ~
        \subfloat[Encoder $E_g$\label{fig:mnist-projection-second-enc}]{%
        \includegraphics[width=1.8in]{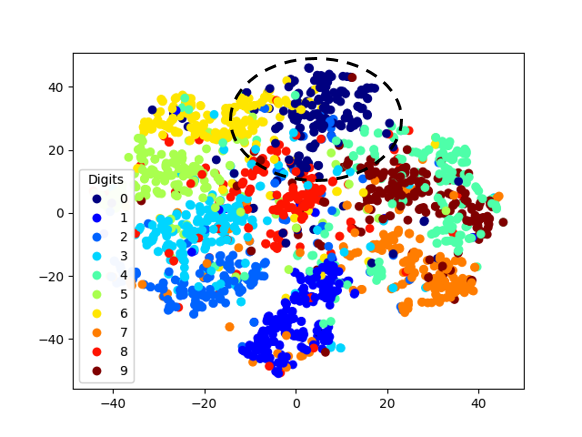}}

    \caption{Latent space projection of encoders in the proposed method on MNIST using abnormal digit 0.}
    \label{fig:mnist-latent-space-projection}
\end{figure}

\begin{figure}
    \centering
    	\subfloat[Encoder $E_r$\label{fig:cifar10-projection-first-enc}]{%
        \includegraphics[width=1.8in]{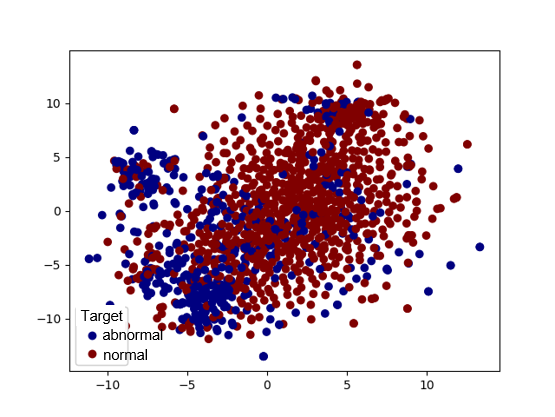}}
	~
		\subfloat[Encoder $E_g$\label{fig:cifar10-projection-second-enc}]{%
        \includegraphics[width=1.8in]{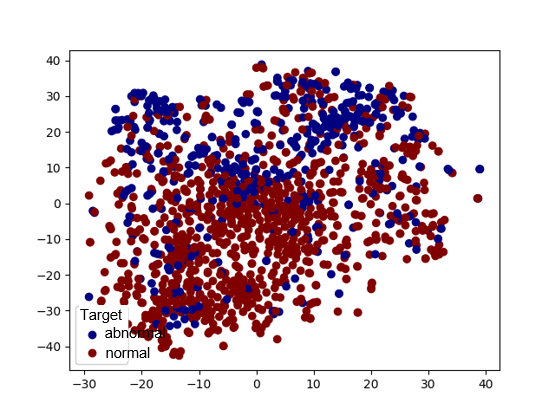}}

    \caption{Latent space projection of encoders in the proposed method on CIFAR10 using abnormal class 'airplane'.}
    \label{fig:cifar10-latent-space-projection}
\end{figure}

In MNIST, we are interested in studying the behavior of the proposed method by projecting the data onto latent space to see the data projection of a model that able to separate normal and abnormal samples. Fig.~\ref{fig:mnist-latent-space-projection} visualizes this data projection of both encoders $E_r$ and $E_g$ on the latent space. According to samples shown for both encoders, the normal and abnormal sample data is distributed to separate the anomaly sample from the whole data. This shows the linearity between the data separation in latent space and data space. When these can be separated in latent space, it might be possible to distinguish between the two groups of samples. For comparison, our argument is supported by the latent space projection of CIFAR10, for which the performance was only AUROC of 0.61 on the data space (see Fig.~\ref{fig:cifar10-latent-space-projection}). 

\section{Conclusion} \label{sec:conclusion}
We proposed an anomaly detection method to reduce bad cycle consistency in BiGAN. This paper assumes that bad cycle consistency might occur due to limitations in the model with respect to preserving enough information from the input image. The proposed method employs a dual-encoder on BiGAN architecture and introduces a preserved information learning mechanism to solve GAN problems as well as to perform anomaly detection. Empirical evaluation on publicly available datasets and brain magnetic resonance imaging anomaly detection showed the performance of the proposed method compared with state-of-the-art methods for separating abnormal samples from the data distribution.

\bibliography{mybib-icmla2020} 
\bibliographystyle{ieeetr}

\end{document}